\documentclass[conference]{IEEEtran}
\IEEEoverridecommandlockouts
\usepackage{amsmath,amssymb,amsfonts}
\usepackage{algorithmic}
\usepackage{graphicx}
\usepackage{textcomp}
\usepackage{xcolor}
\def\BibTeX{{\rm B\kern-.05em{\sc i\kern-.025em b}\kern-.08em
    T\kern-.1667em\lower.7ex\hbox{E}\kern-.125emX}}

\usepackage[
    backend=biber,
    style=ieee,
    sorting=none,% nyvt if by name
    natbib=true,
    url=true,
    doi=true,
    eprint=false,
    maxcitenames=3,
    maxbibnames=99,
    autocite=superscript
]{biblatex}
\begin{document}

\title{The European Moon Rover System: a modular multipurpose rover for future complex lunar missions\\
\thanks{The EMRS rover has been developed during European Moon Rover Sys-tem Pre-Phase A project fully funded by ESA under grant agreement No. 4000137474/22/NL/GLC.}
\thanks{Corresponding authors: Alba Guerra, aguerra@gmv.com; Cristina Luna, cluna@gmv.com}
}

 \author{%
        \IEEEauthorblockN{
            Cristina Luna%
            \IEEEauthorrefmark{1},
            Manuel Esquer%
            \IEEEauthorrefmark{1},
            Jorge Barrientos-Díez%
            \IEEEauthorrefmark{1},
            Alba Guerra%
            \IEEEauthorrefmark{1},
            Marina L. Seoane%
            \IEEEauthorrefmark{1},\\
            Iñaki Colmenarejo%
            \IEEEauthorrefmark{1},
            Steven Kay%
            \IEEEauthorrefmark{2},
            Angus Cameron%
            \IEEEauthorrefmark{2},
            Carmen Camañes%
            \IEEEauthorrefmark{3},
            Íñigo Sard%
            \IEEEauthorrefmark{3},\\
            Danel Juárez%
            \IEEEauthorrefmark{3},
            Alessandro Orlandi%
            \IEEEauthorrefmark{4},
            Federica Angeletti%
            \IEEEauthorrefmark{4},
            Vassilios Papantoniou%
            \IEEEauthorrefmark{5},
            Ares Papantoniou%
            \IEEEauthorrefmark{5},\\
            Spiros Makris%
            \IEEEauthorrefmark{5},
            Armin Wedler%
            \IEEEauthorrefmark{6},
            Bernhard Rebele%
            \IEEEauthorrefmark{6},
            Jennifer Reynolds%
            \IEEEauthorrefmark{7}
            Markus Landgraf%
            \IEEEauthorrefmark{7},
        }\\
        \IEEEauthorblockA{
            \IEEEauthorrefmark{1}
            GMV Aerospace and Defence SAU, Calle de Isaac Newton 11, 
            Tres Cantos, Madrid, Spain, }
            
        \IEEEauthorblockA{
            \IEEEauthorrefmark{2}
            GMV NSL Ltd, Airspeed 2, Eighth Street, Harwell Campus, Oxfordshire, UK, OX11 0RL
            }
            
        \IEEEauthorblockA{
            \IEEEauthorrefmark{3}
            AVS Added Value Solutions, Elgoibar, Spain}
            
        \IEEEauthorblockA{
            \IEEEauthorrefmark{4}
            OHB System AG, Manfred-Fuchs-Strasse 1, 82234 Wessling, Germany}
            
        \IEEEauthorblockA{
            \IEEEauthorrefmark{5}
            Hellenic Technology of Robotics, Kfisias Ave 188, Athens 14562, Greece}
            
        \IEEEauthorblockA{
            \IEEEauthorrefmark{6}
            German Aerospace Center (DLR), Institute of Robotics and Mechatronics, Münchener Str. 20, 82234 Weßling, Germany}

        \IEEEauthorblockA{%
            \IEEEauthorrefmark{7}
            ESTEC, ESA, Keplerlaan 1, 2201 AZ Noordwijk, The Netherlands
        }
    }

\maketitle
\thispagestyle{plain}
\pagestyle{plain}

\begin{abstract}
This document presents the study conducted during the European Moon Rover System Pre-Phase A project, in which we have developed a lunar rover system, with a modular approach, capable of carrying out different missions with different objectives. This includes excavating and transporting over 200kg of regolith, building an astrophysical observatory on the far side of the Moon, placing scientific instrumentation at the lunar south pole, or studying the volcanic history of our satellite. To achieve this, a modular approach has been adopted for the design of the platform in terms of locomotion and mobility, which includes onboard autonomy, of course. A modular platform allows for accommodating different payloads and allocating them in the most advantageous positions for the mission they are going to undertake (for example, having direct access to the lunar surface for the payloads that require it), while also allowing for the relocation of payloads and reconfiguring the rover design itself to perform completely different tasks.
\end{abstract}

\begin{IEEEkeywords}
space robotics, moon rover, emrs, lunar exploration
\end{IEEEkeywords}

\section{Introduction}
Designing a multipurpose rover for future lunar missions requires starting from a modular paradigm and an exhaustive analysis of the lunar environment to understand the key factors that allow developing an optimal mobile solution for the different missions. The European Moon Rover System (EMRS) Pre-Phase A activity fits in the frame-work of the European Exploration Envelope Programme (E3P), which is currently calling for a versatile surface mobility solution to further advance lunar exploration activities \citep{Luna2023ELS}. Future lunar missions are in-deed envisioned to rely on a multimission landing capability, namely the ARGONAUT, formerly the European Large Logistic Lander (EL3), to carry out different science applications. Among them, four missions – currently in pre-Phase A status - require a mobile solution: Polar Explorer (PE), In-Situ Resource Utilisation (ISRU), Astrophysics Lunar Observatory (ALO) and Lunar Geological Exploration Mission (LGEM) \citep{luna2023modularity}.

In this context, the primary aim of this study is to preliminary design the European Moon Rover System (EMRS) in order to achieve modularity and flexibility for use in various mission configurations, while striking a balance between mission versatility and system optimally. Each mission is optimised to fulfil its specific objectives, maximising the utilisation of available resources. This undertaking marks a shift in approach, as the rover must be compatible with four distinct mission scenarios, each with its own independent payload. As a result, this study emphasises a rover concept where flexibility is the driving force. The modularity approach is discussed in relation to the following rover subsystems: rover solution, which includes locomotion, mobility, thermal and space environment studies and autonomous on-board system.

To validate our modular mobility concept, the EMRS solution has already undergone testing at an analogue facility. This paper presents our modular rover solution for future lunar missions, along with the results and insights derived from the testing process.

\section{State-of-the-art}
In the field of robotic mobility systems designed for exploring Mars and the Moon, numerous advanced rover projects are influencing the present forefront of technology. These rovers, either scheduled for imminent deployment or already actively engaged in missions, stand at the vanguard of planetary exploration.

VIPER (NASA): Currently in development, is set to embark on a lunar mission dedicated to exploring the Moon's southern pole in search of frozen water deposits \citep{viper2020}. Scheduled for a late 2024 lunar landing, VIPER boasts a weight of 430 kg, with dimensions measuring 1.53m x 1.53m in footprint and a height of 2.45m \citep{viper2020}. Its scientific instruments include a one-meter-long regolith and ice drill, a Neutron Spectrometer System, a Near-Infrared Volatiles Spectrometer, and a Mass Spectrometer \citep{perseverance2020}. The rover's locomotion system comprises four identical modules, each equipped with wheels, independent steering capabilities, and an active suspension system. These modules are customised for specific locomotion modes, enabling VIPER to navigate low compaction sand effectively, with an average maximum speed of 0.2 m/s \citep{perseverance2020}.

PERSEVERANCE (NASA, 2020): Presently active on Mars, the Perseverance rover is engaged in a multifaceted scientific mission. This includes searching for indications of ancient microbial life, gathering samples of Martian regolith, and conducting groundbreaking experiments such as the historic flight of the helicopter, Ingenuity \citep{marshelicopter} \citep{learnrover}. The rover itself measures 2.9 meters in length, 2.7 meters in width, 2.2 meters in height, and weighs 1025 kg \citep{roboticarm-nasa}. It is equipped with a robotic arm that houses an adaptable drilling system, along with a variety of spectrometers, cameras, environmental sensors, and an ISRU oxygen synthesizer, rendering Perseverance a versatile scientific platform \citep{instruments-nasa},\citep{roverwheels-nasa}. Its six-wheeled chassis incorporates both front and rear wheels with steering capabilities, along with a rocker-boogie suspension system, enabling it to achieve speeds of up to 0.042 m/s

ZHURONG (CNSA): Arriving on Mars in May 2021 \citep{zhurong-cnsa}, the Zhurong rover has a focused mission: to unearth signs of frozen water on Mars while simultaneously analysing the composition of the planet's surface and atmosphere. Measuring 1.23 meters in length, 0.83 meters in width, and 540 meters in height, with a weight of 240 kg \citep{zhuronglocalization}, Zhurong utilises an active boogie-rocker suspension system, providing all six wheels with active steering for adaptable mobility modes \citep{zhurong-additional-cnsa}. It attains a top speed of 0.055 m/s \citep{zhuronglocalization}.

ROSALIND FRANKLIN (ESA): As Europe's pioneering Mars rover, the Rosalind Franklin rover, although fully developed, is currently awaiting its launch \citep{exomars-esa}. Its primary mission revolves around the quest to determine the existence of past life on Mars and involves a suite of instruments, including neutron and infrared spectrometers, an organic molecule analyser, a Raman spectrometer, and a drill capable of reaching depths of 2 meters \citep{exomars-factsheet}. Rosalind Franklin boasts dimensions of 2.5 meters in width, 2 meters in height, and has a weight of 310 kg \citep{exomars-main-esa}. Its mobility is facilitated by six-wheel modules with steering capabilities and a "knee" joint for wheel-walking, all integrated into a 3-boogie passive suspension system \citep{PATEL2010227}. The rover is expected to operate at an average speed of 7.89*104 m/s \citep{exomars-main-esa}.

\section{Multipurpose Rover}
In this section, we will delve into a more comprehensive explanation of the rover platform's mechanical design, as well as the control software and autonomy system architecture.

Five elements are identified within a locomotion system: Mobility, Steering, Suspension, Chassis and Locomotion Software System. 

\subsection{Mobility}
Mobility refers to the propulsion system, and in this case, discrete and hybrid solutions have been disregarded in favour of a continuous mobility solution, specifically wheels. HTR Adaptable Wheels for Exploration (AWE) have been selected for EMRS due to the capacity and maturity of the flexible wheel concept and favourable mass-to-wheel performance ratio of wheeled systems \citep{Iizuka2009, papantoniou2019energetics}.

\begin{figure}[ht]
    \centering
    \includegraphics[scale=0.5]{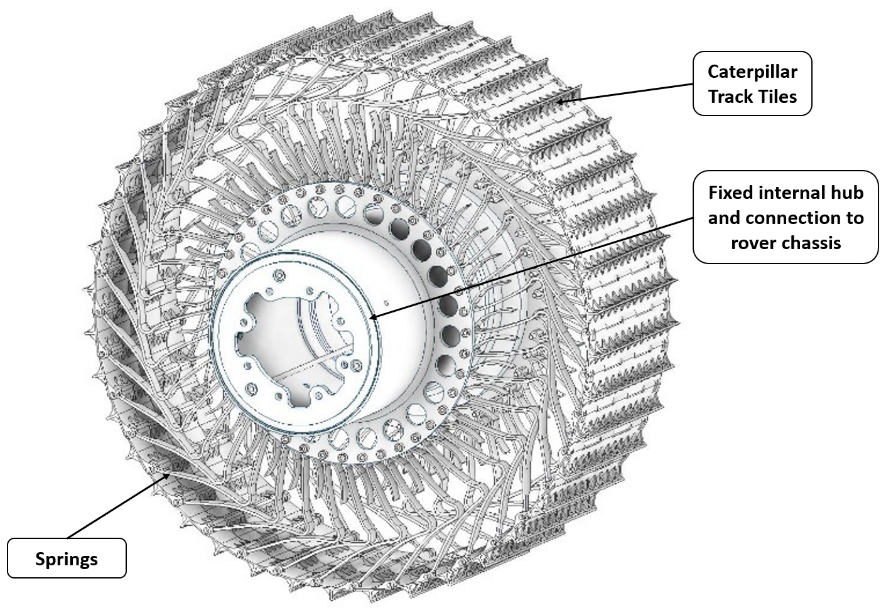}
    \caption{EMRS wheels developed by HTR}
    \label{fig:htr wheel}
\end{figure}

The parameters of the configuration of the AWE (Adaptable Wheels for Exploration) developed by HTR are summarised in the following table:

\begin{table}[h]
       \caption{HTR's AWE parameters for EMRS}
\begin{center}
    \begin{tabular}[width=\columnwidth]{c c}
        \hline
        Parameter & Value/Description\\
        \hline\hline
        Diameter & 612 mm\\ 
        Tile Width & 216 mm\\
        Radial Stiffness & 2500 N/m - 6000 N/m\\
        Max Speed & 3 Km/h \\
        Max torque per wheel & 80 Nm \\  
        Wheel total mass with in hub motor & 7000 g\\
        \hline
    \end{tabular} 
    \label{tab:my_table}
\end{center}
\end{table}

Rover wheels play a pivotal role, especially given the challenging lunar surface conditions and regolith. They must deliver sufficient traction on loosely compacted sand, surmount minor obstacles, and endure the wear and tear over the course of the mission. The wheels designed for this project distinguish themselves through their remarkable flexibility, which can be fine-tuned in the flight model, as well as their utilisation of materials capable of maintaining their mechanical integrity despite the extreme temperature variations experienced on the lunar surface.

\subsection{Steering}

Steering involves rotating the wheel's directions to enable the rover to rotate. In the study two potential options have been identified: on top steering and on side steering. On top steering is typically preferred as it reduces skidding during rotation and the required rotational envelope. However, the on side steering concept is selected because it is more compatible with the upper payload bay volume required for different missions and the stowed envelope.

\begin{figure}[ht]
    \centering
    \includegraphics[width=\columnwidth]{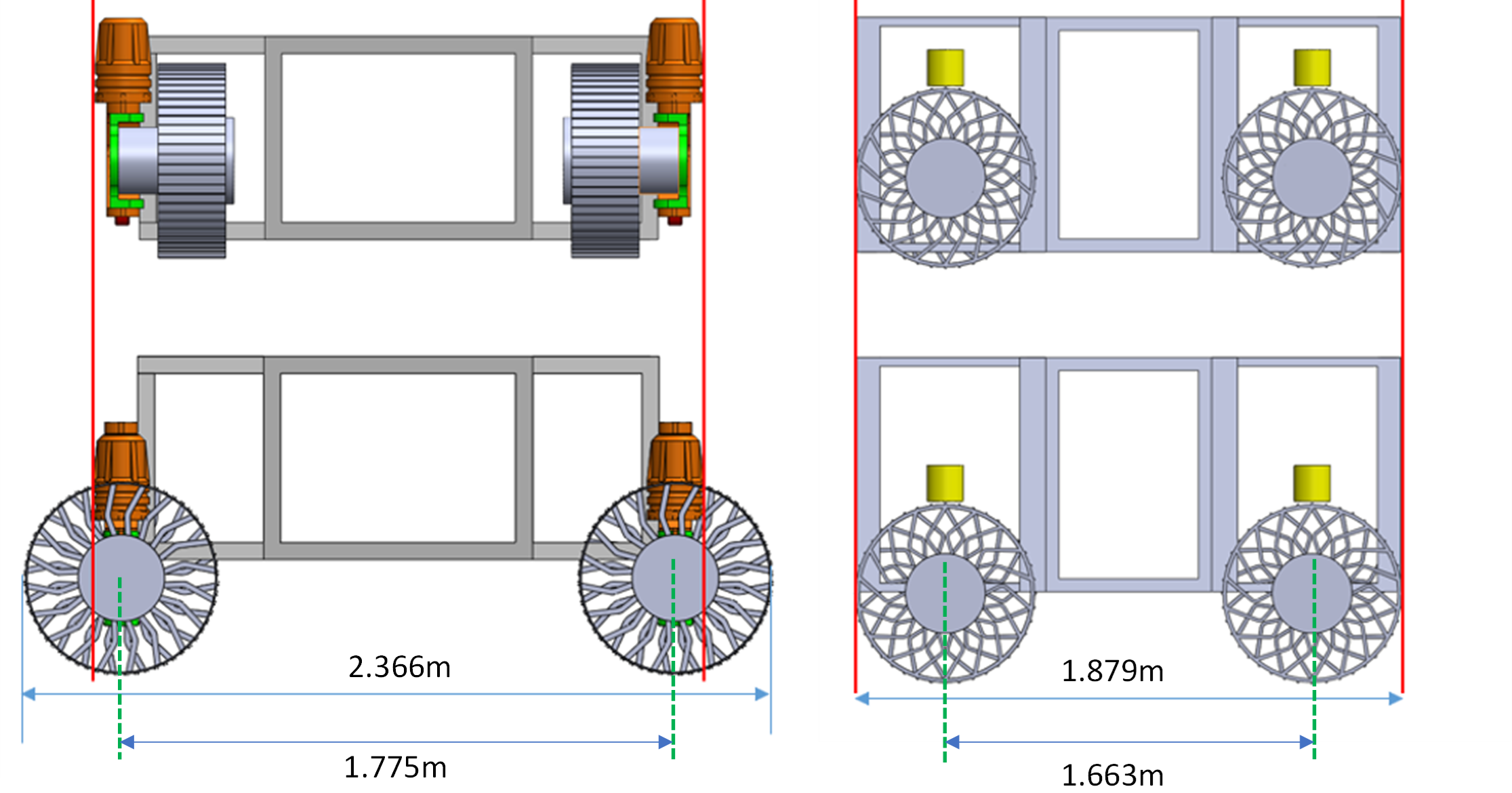}
    \caption{Steering system trade-off}
    \label{fig:steering trade-off}
\end{figure}

\begin{figure}[ht]
    \centering
    \includegraphics[width=\columnwidth]{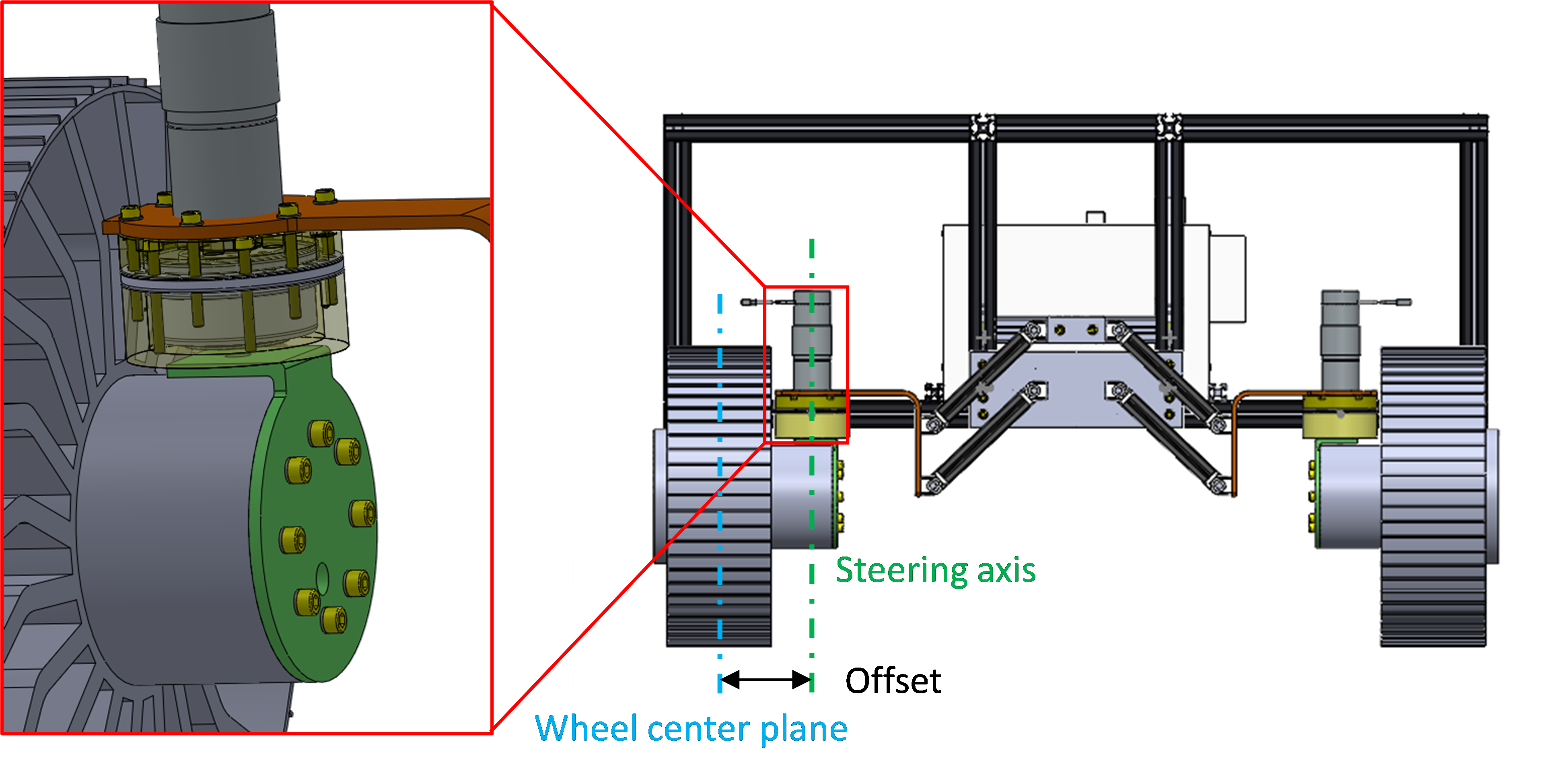}
    \caption{Steering system detail}
    \label{fig:steering detail}
\end{figure}

Having the steering system in offset allows us to gain in manoeuvrability and space. After the functional study with this steering mode, the selected components of the system are: a DC brushed motor with integrated Hall effect incremental encoder, a magnetic detent brake, a planetary gearbox with a harmonic drive, angular ball bearings in back-to-back configuration for the output shaft, spring-energised seal and an absolute output sensor (fig \ref{fig:steering detail}).

The steering actuator is allocated above the hub of the wheel, so that the offset with respect the mid-line of the wheel is relatively small, and mainly driven by wheel width (216 mm) and the steering actuator diameter (150 mm), with some margin for kinematic and mechanical clearances, for a total lever of approximately 200 mm. This leverage increases the bending torques to be transmitted through the steering axis and the required steering output torque compared with the steering-on-top configuration, so the sizing of the components must be adjusted accordingly, but the advantages in payload accommodation are considered worth the additional load levels.

\begin{figure}[ht]
    \centering
    \includegraphics[width=\columnwidth]{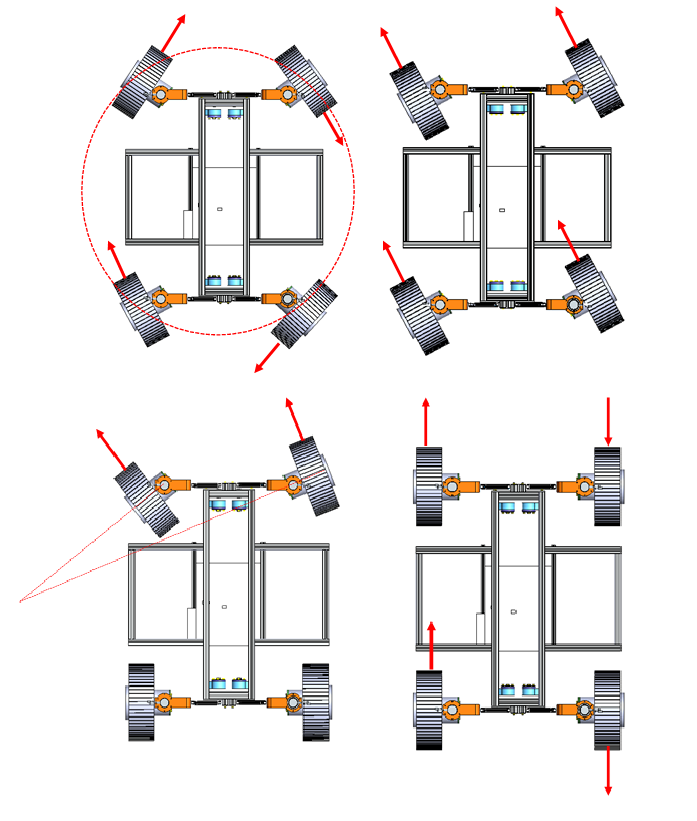}
    \caption{Steering modes}
    \label{fig:steering modes}
\end{figure}

The steering units allow each wheel to change their orientation independently up to $\pm$90 deg. This enhances the rover's mobility capabilities, facilitates movement in narrow terrains where it may be more advantageous to switch between different modes of locomotion. The possible locomotion modes for the rover include Skid Steering, Ackermann Turn, Crab Turn, and Point Turn (fig \ref{fig:steering modes}). The system enables the alteration of locomotion modes at any point during the traversal, improving mobility and enabling more effective obstacle avoidance. Furthermore, turning the wheels up to 90 degrees allows for folding them, reducing the size of the footprint and enabling the dimensions to conform to those required by the multi-mission landing capability, the European Large Logistic Lander (EL3).

\subsection{Suspension}
Suspension aims to compensate for the loads between the wheels during traversal, ensuring traction in challenging terrains such as overcoming rocks. It is particularly crucial on the Moon, due to low gravity and variable terrain compactness, which can risk stability during traversal. The study identifies passive and active suspension solutions from the literature. The selected solution is a hybrid one, with a baseline of a passive suspension based on a parallelogram articulated fork with preloaded elements. 

The suspension system is composed of an independent suspension for each of the 4 wheels. This is based on a linkage system to compensate the offset to actual wheel contact region. The linkages have been implemented in the frontal planes since they allow the most advantageous stowed configuration. The use of independent suspension assemblies for each wheel enables also for a compact stowage while the active nature of the solution assisting to the deployment.

The suspensions bars feature the same length and have their supports vertically aligned, so that the wheels are kept nominally horizontal. These ones are made of CFRP tubes with Ti6Al4V end fittings containing the corresponding dust-sealed rotary joints.

\begin{figure}[h!]
    \centering
    \includegraphics[width=\columnwidth]{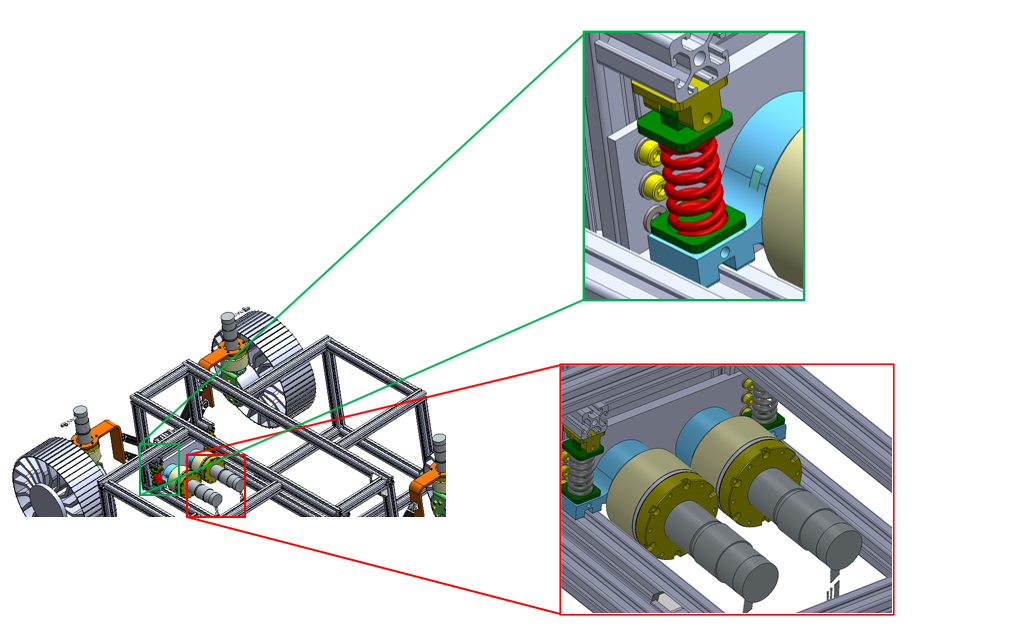}
    \caption{Passive and active suspension}
    \label{fig:Passive and active suspension}
\end{figure}

Additionally, an option is included to implement an active suspension system on top of the passive suspension, which would have minimal impact on the rest of the rover platform. This option is reserved for missions with challenging use cases, such as ISRU missions that involve carrying up to 200kg of regolith, which can impact the stability of the rover. The active suspension allows modification of the centre of gravity when the rover is fully loaded to ensure stability.

The suspension features an active system in series with a passive system, so that the actuators are ‘floating’ regulating the relative angle between the lever attached to the passive spring-driven suspension and the output shaft. The following schematic shows the working principle:

\begin{figure}[h!]
    \centering
    \includegraphics[width=\columnwidth]{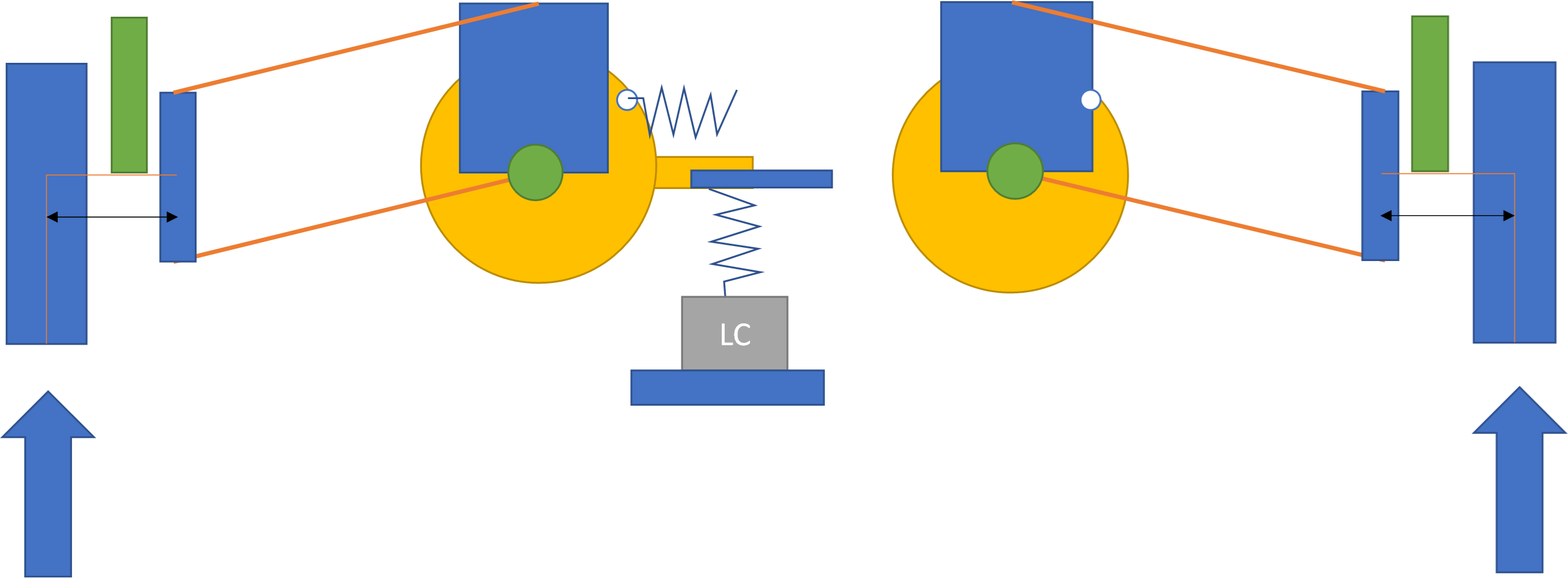}
    \caption{In-series active/passive suspension working principle}
    \label{fig:suspension working principle}
\end{figure}

The independent active suspension enables a wheel-walking mode, and coupled with the independent steering even a ’paddling’ mode, which while not being part of the locomotion baseline, constitute a very interesting safeguard in case of getting stuck in very soft terrain, which is one of the most challenging scenarios for conventional passive suspension systems. Furthermore, this suspension concept allows direct access to the ground for payloads that require it, such as drills or spectrometers.

As mentioned earlier, during the design phase, saving space in Argonaut has been considered a key factor. Therefore, the mobility system as a whole allows for a "stowed" mode, retracting the wheels and lowering the suspension, significantly reducing the space occupied by the rover:

\begin{figure}[ht]
    \centering
    \includegraphics[width=\columnwidth]{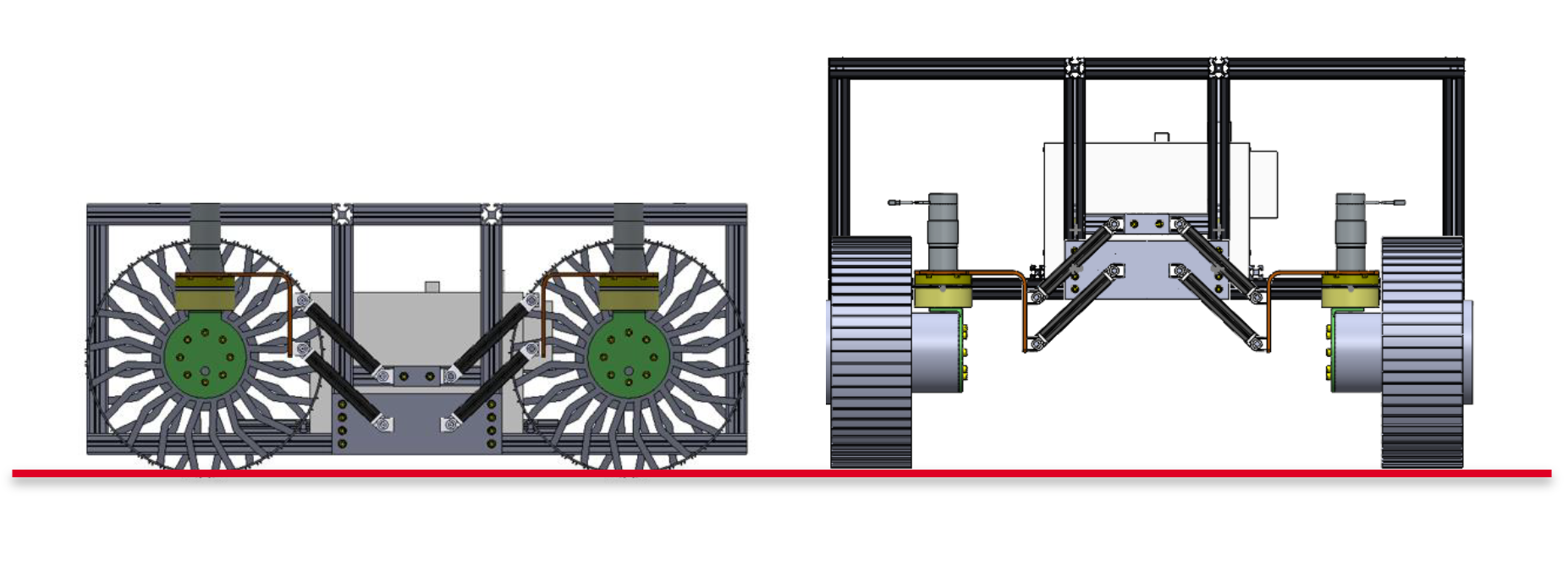}
    \caption{Stowed and normal configuration}
    \label{fig:stowed}
\end{figure}

\subsection{Chassis}
\begin{figure}[ht]
    \centering
    \includegraphics[scale=0.1]{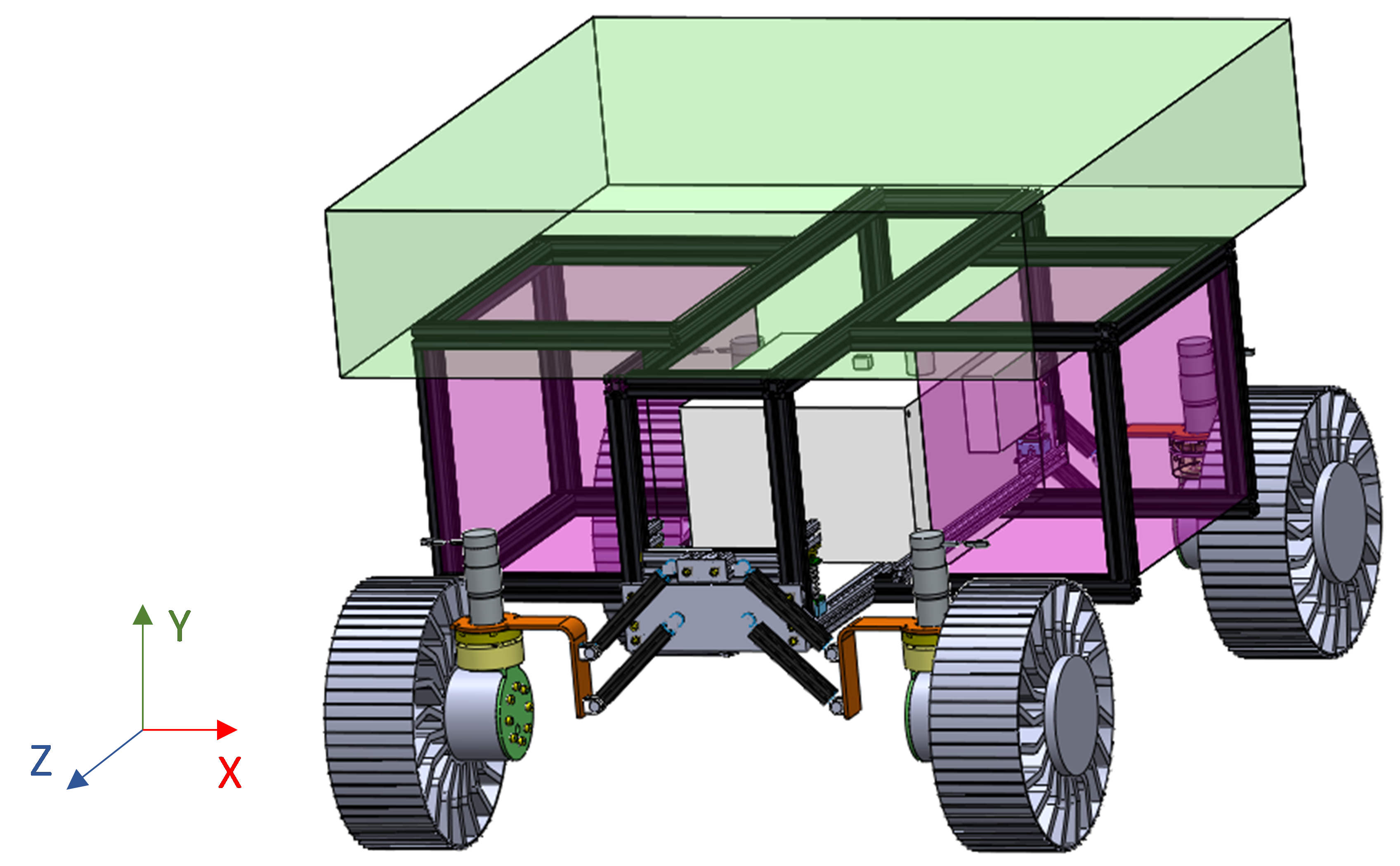}
    \caption{Proposed solution with the modular side payload volumes (pink) and top modular payload volume (green)}
    \label{fig:chassis solution}
\end{figure}

The proposed chassis solution is a modular structure constructed from CFRP (Carbon Fiber Reinforced Polymer). This modularity is based on the versatility of the locomotion system and the compartmentalisation of various rover compartments to accommodate diverse payloads and subsystems, as illustrated in the figure \ref{fig:chassis solution}.

The primary bay, centrally located at the bottom of the rover and shared across all missions, serves to distribute the rover's loads to the locomotion system and houses common avionics, thermal control systems, and power components. Flanking this central bay, lateral payload modules make efficient use of the space between the locomotion working envelopes to carry payloads, with a priority on those requiring direct access to the lunar surface, such as drills/excavators or GPRs. Additionally, at each side of the main receptacle, another payload volume is available to incorporate instruments like robotic arms, camera masts, or antennas, among others. These two configurable payload volumes can be adapted for different missions, allowing for the reuse of the remaining rover elements.

\begin{figure}[h]
    \centering
    \includegraphics[width=\columnwidth]{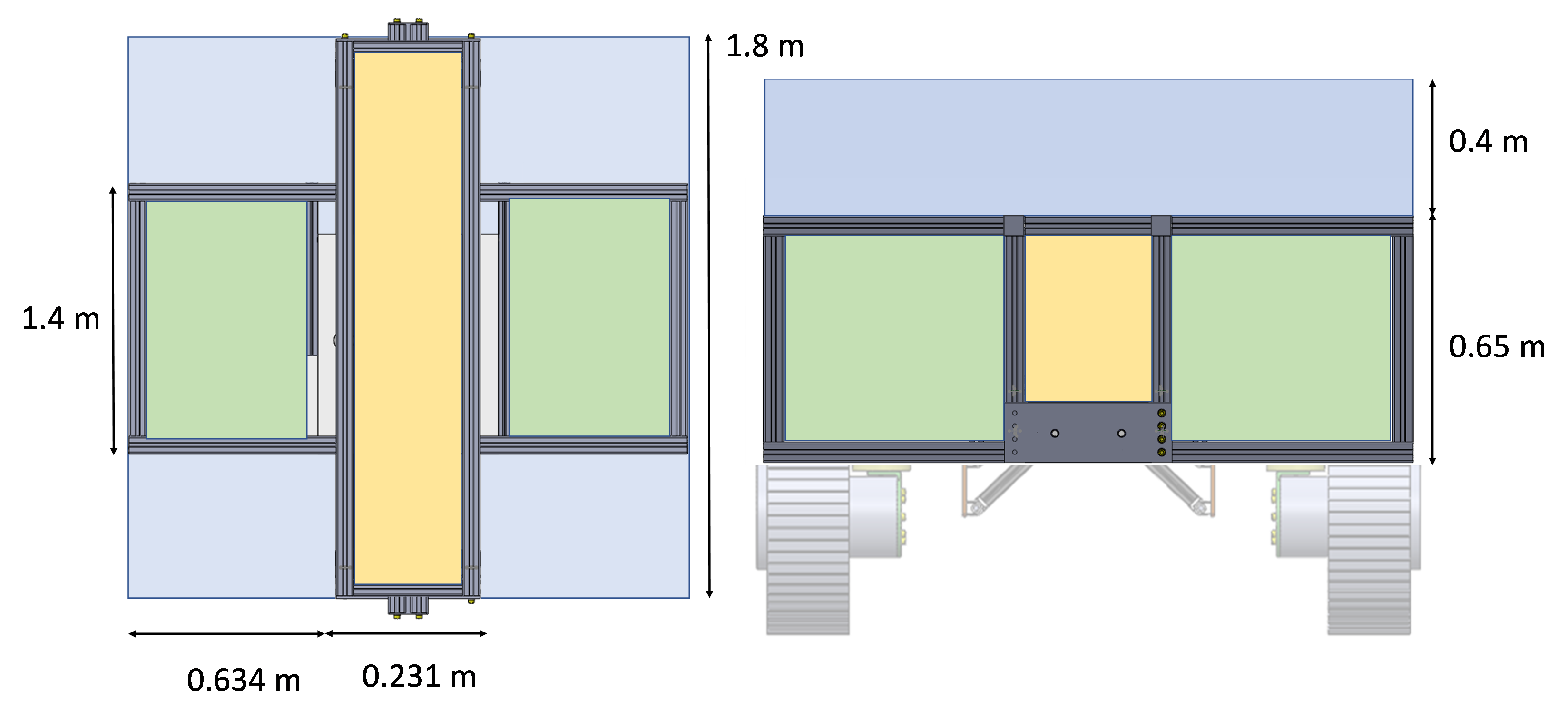}
    \caption{Chassis dimensions}
    \label{fig:chassis dimensions}
\end{figure}

The overall chassis dimensions are summarised in the following table:

\begin{table}[ht]
       \caption{Module dimensions summary}
\begin{center}
    \begin{tabular}{c c c c c}
        \hline
        Module & Length ($m$) & Width ($m$) & Height ($m$) & Total volume ($m^3$)\\
        \hline\hline
        Central & 1.8 & 0.231 & 0.65 & 0.25\\ 
        Lateral  & 1.4 & 0.634 & 0.4 & 0.34\\ 
        Top & 1.8 & 1.5 & 0.4 & 1.12\\ 
        \hline
    \end{tabular} 
    \label{tab:chassis dimensions}
\end{center}
\end{table}

\subsection{Locomotion Software System}
This software architecture has been developed to manage the rover's degrees of freedom, enabling it to navigate using various locomotion modes. The flow chart of the system is described in figure 

\begin{figure}[h]
    \centering
    \includegraphics[width=\columnwidth]{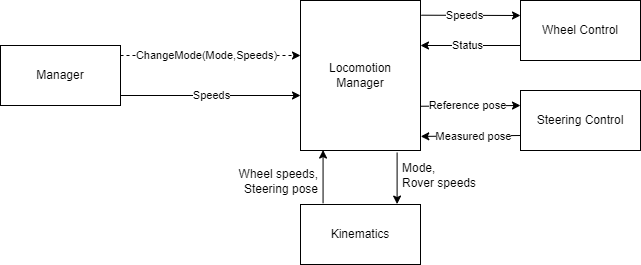}
    \caption{Block diagram of the locomotion software implemented in the avionics.}
    \label{fig:software_architecture}
\end{figure}

The different modules have the following characteristics:
\begin{enumerate}
    \item \textbf{Manager:} An interface module is designed to receive various system commands, which are categorised into speed adjustment and kinematic mode change commands. These inputs are configured to be controlled using a joystick controller.
    \item \textbf{Locomotion Manager:}This module is responsible for synchronising the flow of the system, acquiring desired set points from the kinematics, and generating trajectory commands for the steering and wheel control modules. It also oversees the execution of locomotion mode changes and ensures the proper functioning of the entire system.
    \item \textbf{Kinematics:} The purpose of this module is to translate vehicle speed commands into motor commands and perform the reverse conversion to calculate locomotion odometry. Considering the current locomotion mode of the rover, generates the corresponding conversions.
    \item \textbf{Wheel Control:} Module designed for implementing a velocity control loop for the wheel motors and conducting a security analysis of their performance.
    \item \textbf{Steering Control:}" Module designed for executing a position control loop for the steering motors, along with a security analysis of their behaviour.
\end{enumerate}

\subsection{Thermal}

The target landing points for EMRS are the following: Shoemaker / Faustini craters in the case of PE; Tsiolkovsky / VonKárman craters located in the farside of the Moon in the case of ALO; Schrödinger crater in the case of ISRU and Ina-D irregular Mare Patch / Rima Bode Region / Imbrium Basin in the case of LGEM. These locations exhibit different illuminations and temperatures, as well as varying types of regolith and flat regions. This not only affects thermal analysis but also dust protection systems and the mobility configuration itself. Therefore, a comprehensive analysis of the lunar environment has been conducted to adapt the common systems for all missions to the selected areas, ensuring the proper performance of the rover in each mission. The study of these analysis has been considered in the design of this project but it will not be explained in this paper. The details will be exposed in future works.

\section{Multipurpose OBSW}

Autonomy is the key for the present and future robotic space and terrestrial applications. The key for autonomy is to provide systems with the capability to make decisions autonomously \citep{CISRU2022}, and more especially if this system is focused on carrying out different missions with different objectives and subsystems. Designing a rover for different missions involves developing software capable of abstracting from specific missions, which implies designing a goal-oriented autonomous system. This is done by providing the rover with dynamic re-planning capability, semantic segmentation for the detection and identification of different obstacles and objects with different shapes, sizes, and weights. It also has manipulation capabilities, allowing it to perform ISRU tasks, manipulate tools, and scientific sampling. Further details of the multipurpose OBSW are discussed in \citep{Luna2023ELS, CISRU2022, romero2023enabling, luna2023modularity}.

\section{EMRS Breadboard}
We have developed an EMRS breadboard to be able to demonstrate our modular mobility concept and to validate the feasibility of having one multipurpose rover for different lunar missions. The main focus is elaborate a versatile rover which satisfy the requirements for the activities included in European Exploration Envelope Programme (E3P) such as Polar Explorer (PE), In-Situ Resources Utilization (ISRU), Astrophysics Lunar Observatory (ALO) and Lunar Geological Exploration Mission (LGEM).

\begin{figure}[ht]
    \centering
    \includegraphics[width=\columnwidth]{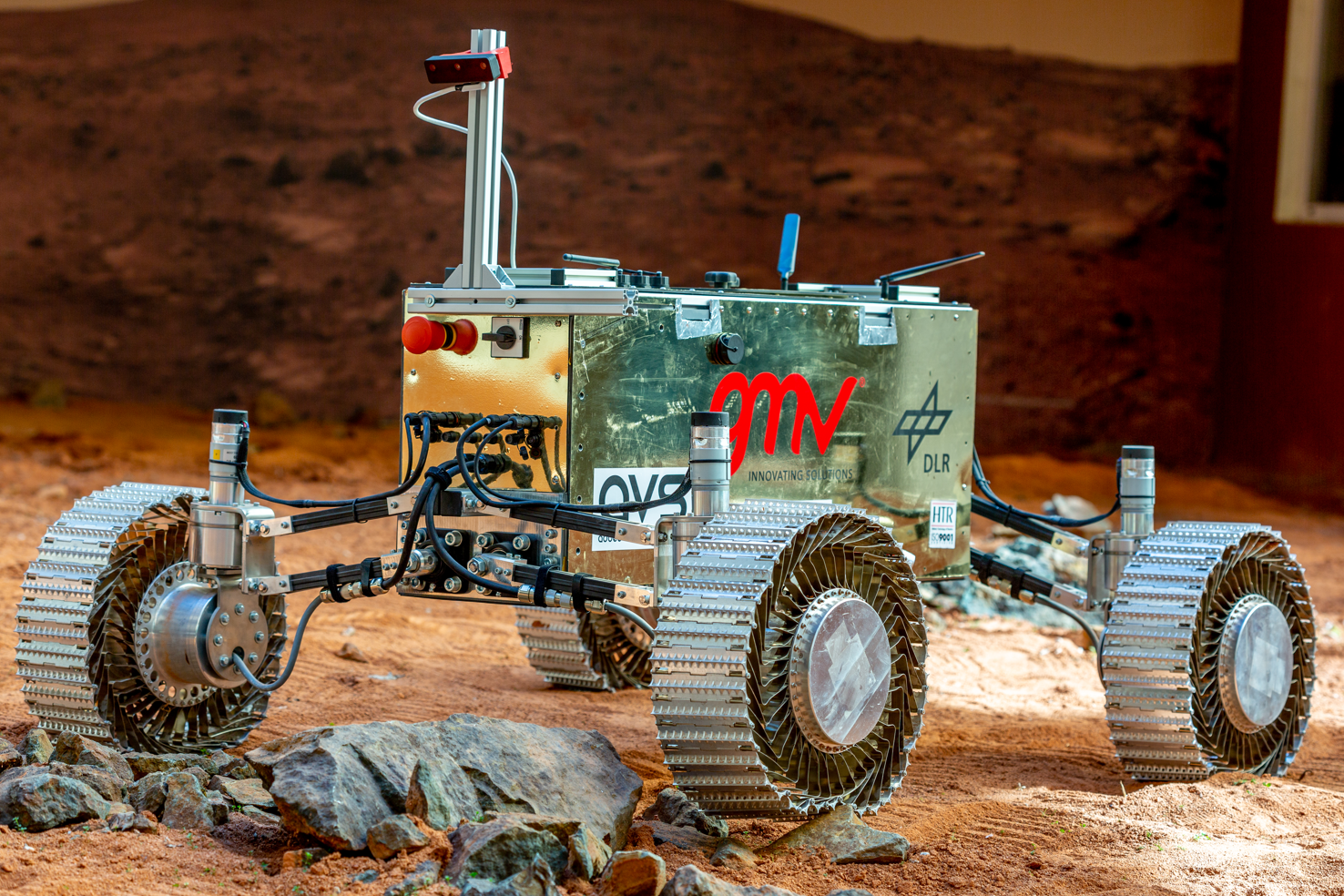}
    \caption{EMRS breadboard GMV SPoT}
    \label{fig:emrsspot}
\end{figure}

One of the main focus of the rover prototype is to demonstrate the locomotive performance in a representative terrain environment. To carry out the development of this rover, it has been executed in a scaled manner in the proportion presented below:

 \begin{align}
  \begin{aligned}
    L_{earth} = L_{moon}·(\rho_{moon}\rho_{earth})^{1/3}\\
    L_{moon} · 0.55 = L_{moon}/1.8
  \end{aligned}
 \end{align}

The result is a scaled rover with the following dimensions:

\begin{enumerate}
    \item \textbf{Dimensions}(L,W,H): 1.879,1.5,0.7 stowed - 2.366, 1.525, 1 deployed
    \item \textbf{Ground clearance:} 0.3 m
    \item \textbf{Distance between wheels:} Longitudinal = 1.775 m, Transversal = 1.284 m
    \item \textbf{Wheel dimensions:} Diameter = 612 mm, Width: 216 mm 
\end{enumerate}

The result is a full operative rover developed using COTS and space components that satisfies all the design requirements (fig \ref{fig:emrsspot}. The breadboard has been tested on representative analogue lunar environment to validate our modular mobility concept for lunar exploration with promising results that are discused in the following section.

\section{Analogue Test Campaign}

The breadboard has been tested during an analogue test campaign. To achieve this, the rover and locomotion concept have primarily been tested by temporarily removing the side and upper modules for logistical reasons during robot transportation.

\begin{figure}[ht]
    \centering
    \includegraphics[width=\columnwidth]{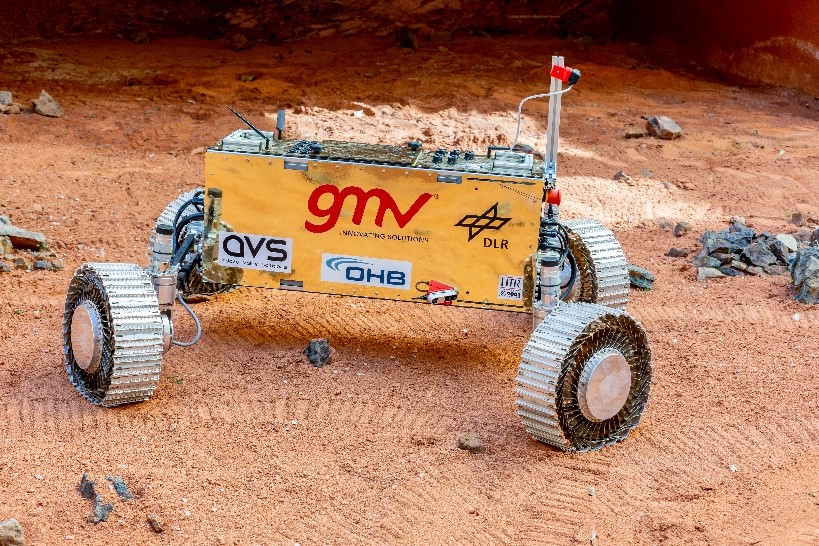}
    \caption{EMRS steering tests at GMV SPoT}
    \label{fig:emrspointturn}
\end{figure}

The field test campaign has been conducted in two phases: phase one at the lunar analogue facility, DLR's Planetary Exploration Laboratory (PEL), and phase two at GMV's Mars SPoT facility. These tests have served as a means to demonstrate the effectiveness of our flexible and modular rover concept. During phase one, we had the opportunity to test the solution in a lunar soil simulant testbed, which provided a realistic environment for assessing the rover's performance. Additionally, phase two allowed us to further evaluate the rover's capabilities by testing various locomotion modes in rocky and uneven terrain.

\begin{figure}[ht]
    \centering
    \includegraphics[width=\columnwidth]{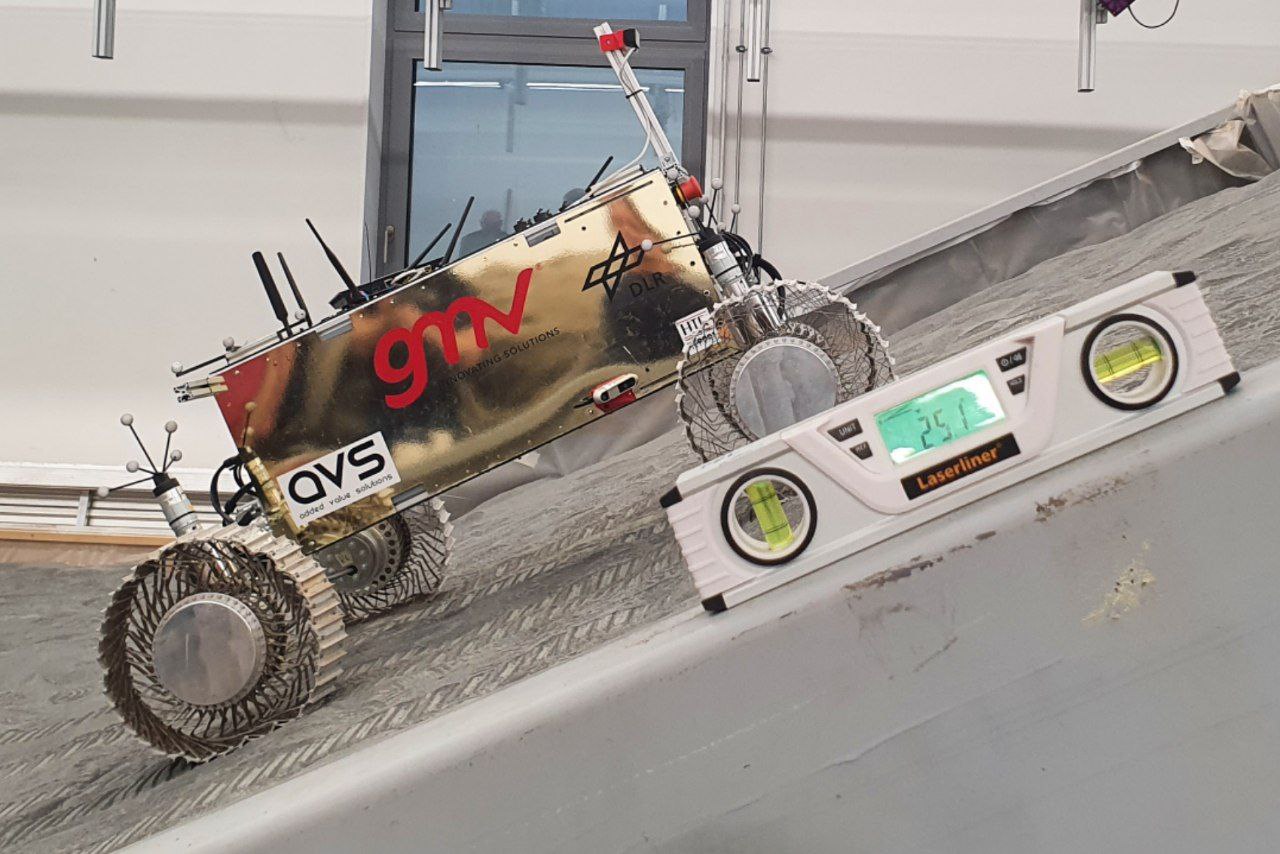}
    \caption{EMRS tests, 25º slope, at DLR PEL}
    \label{fig:emrs25}
\end{figure}

Various tests have been conducted with different terrain inclinations, surpassing inclines of up to 25 degrees (fig \ref{fig:emrs25}). This solidly meets and even exceeds the requirements, demonstrating the crater-traversing capability of our design.

\begin{figure}[ht]
    \centering
    \includegraphics[width=\columnwidth]{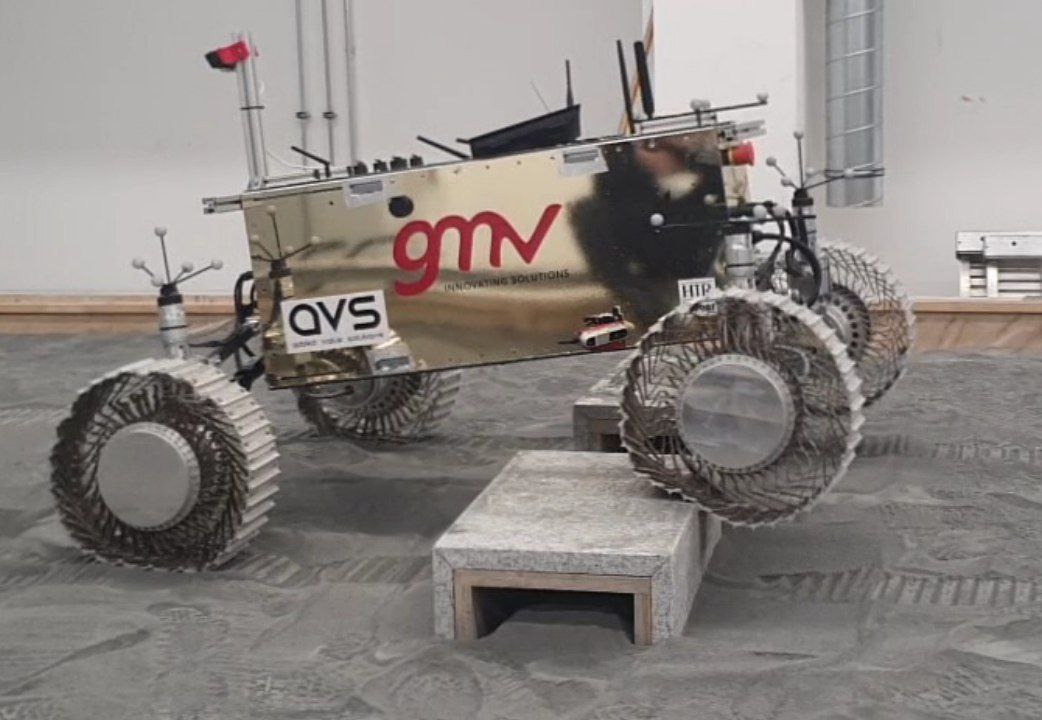}
    \caption{EMRS tests, 30cm obstacle overcoming, at DLR PEL}
    \label{fig:emrs30}
\end{figure}

Another key objective was to demonstrate the rover's ability to traverse rocky or obstacle-laden environments. Therefore, tests were conducted with both natural and artificial obstacles at various heights (see Figure \ref{fig:emrs30}). In this regard, the suspension and wheel systems have enabled the rover to navigate obstacles of up to 30 cm without any issues.

\begin{figure}[h]
    \centering
    \includegraphics[width=\columnwidth]{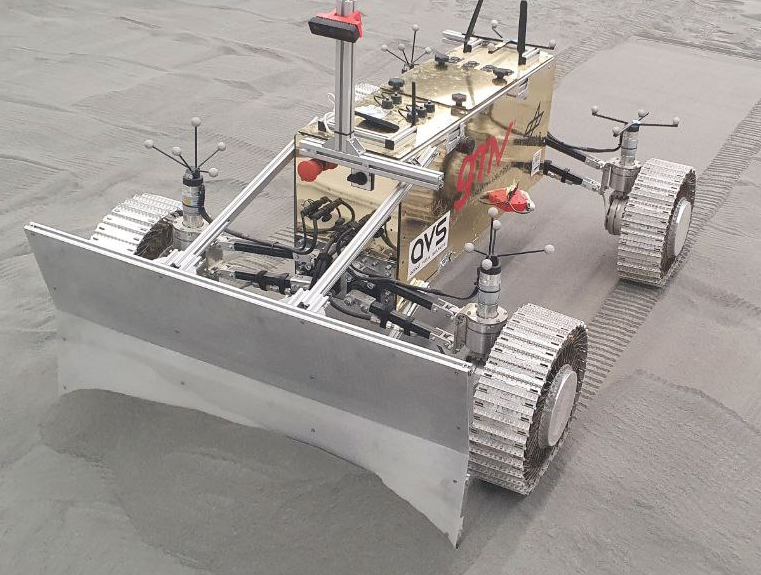}
    \caption{EMRS tests, excavator, at DLR PEL}
    \label{fig:emrsex}
\end{figure}

Excavating lunar regolith is one of the most challenging tasks that EMRS will have to undertake. Therefore, utilising the DLR PEL lunar simulant, we conducted excavation tests to verify that the rover's locomotion and mobility system was capable of generating the required traction while excavating regolith (see Figure \ref{fig:emrsex}).

These tests have provided valuable insights into the practical implementation of our rover design and have further validated the feasibility of our modular approach.

\section{Conclusions}
This paper presents the concept of a versatile modular rover, developed as part of the EMRS project. Throughout this project, not only has the rover been conceptualised, but critical systems related to power management, thermal control, insulation, and dust protection have also been designed. It is noteworthy that these systems will be the focal point of upcoming investigations and refinements.

A series of rigorous obstacle and excavation tests has been carried out, shedding light on the remarkable capabilities of the EMRS system configuration. These tests not only demonstrate the rover's ability to safely navigate lunar terrain but also underscore its proficiency in lunar regolith excavation. The latter is of paramount importance in establishing the groundwork for sustainable human habitation on the Moon, as it enables the extraction of essential resources required for sustaining life and advancing scientific exploration.

The modular design of our prototype provides a unique advantage by allowing us to evaluate the rover's locomotion and software in tandem with a variety of scientific payloads. These payloads include diverse instruments such as neutron spectrometers, drills, and various cameras. This adaptability empowers us to explore and optimise the rover's performance across a range of scientific objectives, thereby maximising its utility in future lunar missions.

\section*{Acknowledgments}

EMRS is nothing more than the success of cooperation among a consortium of European companies that, through ESA, can demonstrate the potential of the space technology we develop. In this regard, it is essential to thank the teams from GMV, AVS, HTR, OHB, and DLR for their cooperation and enthusiasm demonstrated throughout the project, which has led to the creation of a rover capable of carrying out many different space missions. We would also like to express our gratitude to ESA for its strong commitment to the European space industry and for the support and resources we have received.

\printbibliography

@inproceedings{Luna2023ELS,
author = {Luna, Cristina and Guerra, Alba and Barrientos, Jorge and Esquer, Manuel and Colmenarejo, Iñaki and Camañes, Carmen and Sard, Iñigo and Juárez, Danel and Arregui, Íñigo and Reynolds, Jennifer},
year = {2023},
month = {06},
title = {DESIGNING A MODULAR MULTIPURPOSE ROVER FOR FUTURE LUNAR MISSIONS},
booktitle    = {11th European Lunar Symposium },
address      = {Padova, Italy},
date         = {27-29 June 2023}

}

@inproceedings{luna2023modularity,
  author       = {Cristina Luna and Jorge Barrientos-Díez and Manuel Esquer and Alba Guerra and Marina López-Seoane and Iñaki Colmenarejo and Fernando Gandía and Steven Kay and Angus Cameron and Carmen Camañes and Íñigo Sard and Danel Juárez and Alessandro Orlandi and Federica Angeletti and Vassilios Papantoniou and Ares Papantoniou and Spiros Makris and Bernhard Rebele and Armin Wedler and Jennifer Reynolds and Markus Landgraf},
  title        = {Modularity for lunar exploration: European Moon Rover System Pre-Phase A Design and Field Test Campaign Results},
  year         = {2023},
  month        = {10},
  booktitle    = {74th International Astronautical Congress (IAC)},
  address      = {Baku, Azerbaijan},
  date         = {2-6 October 2023}
}

@inproceedings{CISRU2022,
    author = {Ocón, Jorge and Dragomir, Iulia and Luna, Cristina and Gandía, Fernando and Estremera, Joaquín},
    year = {2022},
    month = {06},
    pages = {},
    title = {Autonomous Decision Making: a key for future robotic applications},
    booktitle = {16th Symposium on Advanced Space Technologies in Robotics and Automation (ASTRA 2022)}
}

@inproceedings{romero2023enabling,
  title={Enabling In-Situ Resources Utilisation by leveraging collaborative robotics and astronaut-robot interaction},
  author={Romero-Azpitarte, Silvia and Luna, Cristina and Guerra, Alba and Alonso, Mercedes and Romeo Manrique, Pablo and Seoane, Marina L. and Olayo, Daniel and Moreno, Almudena and Castellanos, Pablo and Gandía, Fernando and Visentin, Gianfranco},
  booktitle={74th International Astronautical Congress (IAC)},
  year={2023},
  month={October},
  address={Baku, Azerbaijan},
  days={2-6},
}

@misc{viper2020,
  title={Viper: The Rover and its Onboard Toolkit},
  author={R. Chen},
  year={2020},
}

@misc{perseverance2020,
   author = {NASA},
   title = {Mars 2020 Perseverance Rover},
   url = {https://mars.nasa.gov/mars2020/},
   year = {2020}
}

@misc{marshelicopter,
   author = {NASA},
   title = {Mars Helicopter},
   url = {https://mars.nasa.gov/technology/helicopter/},
}

@article{PATEL2010227,
author = {Patel, Nildeep and Slade, Richard and Clemmet, Jim},
doi = {https://doi.org/10.1016/j.jterra.2010.02.004},
issn = {0022-4898},
journal = {Journal of Terramechanics},
number = {4},
pages = {227--242},
title = {{The ExoMars rover locomotion subsystem}},
url = {https://www.sciencedirect.com/science/article/pii/S0022489810000182},
volume = {47},
year = {2010}
}

@unknown{zhuronglocalization,
author = {Zhou, Ruyi and Yu, Tianyi and Gao, Haibo and Yang, Huaiguang and Li, Jian and Yuan, Ye and Liu, Chuankai and Wang, Jia and Zhao, Yuyan and Wang, Zhengyin and Wang, Xiyu and Bao, Gang and Deng, Zongquan and Huang, Lan and Li, Nan and Cui, Xiaofeng and He, Ximing and Jia, Yang and Di, Kaichang},
doi = {10.21203/rs.3.rs-836162/v1},
title = {{Localization and Surface Characterization by Zhurong Mars Rover at Utopia Planitia}},
year = {2021}
}

@online{exomars-main-esa,
author = {ESA},
title = {{ExoMars}},
url = {https://www.esa.int/Science{\_}Exploration/Human{\_}and{\_}Robotic{\_}Exploration/Exploration/ExoMars}
}

@online{exomars-factsheet,
author = {ESA},
title = {{ExoMars factsheet}},
url = {https://www.esa.int/Science{\_}Exploration/Human{\_}and{\_}Robotic{\_}Exploration/Exploration/ExoMars/ExoMars{\_}Factsheet}
}

@online{zhurong-additional-cnsa,
author = {CNSA},
title = {{Zhurong Rover Additional Information}},
url = {http://www.cnsa.gov.cn/english/n6465652/n6465653/c6811942/content.html}
}

@online{instruments-nasa,
author = {NASA},
title = {{Instruments}},
url = {https://mars.nasa.gov/mars2020/spacecraft/instruments/}
}

@online{roverwheels-nasa,
author = {NASA},
title = {{Rover Wheels}},
url = {https://mars.nasa.gov/mars2020/spacecraft/rover/wheels/}
}

@online{zhurong-cnsa,
author = {CNSA},
title = {{Zhurong Rover Information}},
url = {http://www.cnsa.gov.cn/english/n6465652/n6465653/c6812005/content.html}
}

@online{exomars-esa,
author = {ESA},
title = {{Rover ready next steps for ExoMars}},
url = {https://www.esa.int/Science_Exploration/Human_and_Robotic_Exploration/Exploration/ExoMars/Rover_ready_next_steps_for_ExoMars}
}

@online{roboticarm-nasa,
author = {NASA},
title = {{Robotic arm}},
url = {https://mars.nasa.gov/mars2020/spacecraft/rover/arm/}
}

@online{learnrover,
author = {NASA},
title = {{Learn about the Rover}},
url = {https://mars.nasa.gov/mars2020/spacecraft/rover/}
}

@article{Iizuka2009,
author = {Iizuka, Kojiro and Kubota, Takashi},
year = {2009},
month = {01},
pages = {},
title = {Study of Flexible Wheels for Lunar Exploration Rovers: Running Performance of Flexible Wheels with Various Amount of Deflection},
volume = {7},
journal = {Journal of Asian Electric Vehicles},
doi = {10.4130/jaev.7.1319}
}

@inproceedings{papantoniou2019energetics,
  author       = {Vassilios Papantoniou and Stylianos Skevakis and Anastasios G. Katelouzos and Ares Papantoniou and Konstantinos Kapellos and Evangelos Papadopoulos and Michel Van Winnendael},
  title        = {Energetics impact from the use of flexible and adaptable stiffness wheels on lunar and planetary rovers},
  year         = {2019},
  booktitle    = {Symposium on Advanced Space Technologies in Robotics and Automation (ASTRA 2019)}
}

\end{document}